\documentclass{article}

\usepackage{arxiv}

\usepackage[utf8]{inputenc} % allow utf-8 input
\usepackage[T1]{fontenc}    % use 8-bit T1 fonts
\usepackage{hyperref}       % hyperlinks
\usepackage{url}            % simple URL typesetting
\usepackage{booktabs}       % professional-quality tables
\usepackage{amsfonts}       % blackboard math symbols
\usepackage{nicefrac}       % compact symbols for 1/2, etc.
\usepackage{microtype}      % microtypography
\usepackage{lipsum}
\usepackage{graphicx}
\graphicspath{ {./images/} }
\usepackage{amsmath}
\usepackage{algorithm}
\usepackage{algorithmic}
\usepackage{subcaption}

\usepackage{array}
\title{HySim-LLM: Embedding-Weighted Fine-Tuning Bounds and Manifold Denoising for Domain-Adapted LLMs}
\author{
Majid Jaberi-Douraki\textsuperscript{1,2,}\thanks{Corresponding Author: jaberi@k-state.edu},  
Hossein Sholehrasa\textsuperscript{1,3}, 
Xuan Xu\textsuperscript{1,4}, 
Remya Ampadi Ramachandran\textsuperscript{1,2} \\[6pt]
\textsuperscript{1}1DATA Consortium and FARAD Program, Kansas State University, Olathe, KS, USA\\
\textsuperscript{2}Department of Mathematics, Kansas State University, Olathe, KS, USA\\
\textsuperscript{3}Department of Computer Science, Kansas State University, Manhattan, KS, USA\\
\textsuperscript{4}Department of Statistics, Kansas State University, Olathe, KS, USA
}

\begin{document}
\maketitle
\vspace{-1em}

\begin{abstract}
The extraction and standardization of pharmacokinetic (PK) information from scientific literature remain significant challenges in computational pharmacology, which limits the reliability of data-driven models in drug development. Large language models (LLMs) have achieved remarkable progress in text understanding and reasoning, yet their adaptation to structured biomedical data, such as PK tables, remains constrained by heterogeneity, noise, and domain shift. To address these limitations, we propose HySim-LLM, a unified mathematical and computational framework that integrates embedding-weighted fine-tuning and manifold-aware denoising to enhance the robustness and interpretability of LLMs. We establish two theoretical results: (1) a similarity-weighted generalization bound that quantifies adaptation performance under embedding divergence, and (2) a manifold-based denoising guarantee that bounds loss contributions from noisy or off-manifold samples. These theorems provide a principled foundation for fine-tuning LLMs in structured biomedical settings. The framework offers a mathematically grounded pathway toward reliable and interpretable LLM adaptation for biomedical and data-intensive scientific domains.
\end{abstract}
% keywords can be removed
%\keywords{First keyword \and Second keyword \and More}

\section{Introduction}
The extraction and standardization of pharmacokinetic (PK) information from scientific literature remain significant bottlenecks in computational pharmacology \cite{riviere2018veterinary}. Due to the absence of a comprehensive, centralized, and up-to-date PK database, researchers must rely on previously published studies to collect and interpret PK parameters \cite{maunsell2021farad}. However, these data are often dispersed across heterogeneous sources, presented in varying formats, and embedded within complex tables or supplementary materials \cite{jaberi2021large}. This heterogeneity makes automatic extraction difficult and prone to errors, thereby reducing the reliability of downstream modeling and analysis. The challenges of identifying, curating, and normalizing PK data thus pose a significant constraint to developing robust algorithms for preclinical and clinical drug development.

Recent advances in large language models (LLMs) have revolutionized natural language processing, enabling state-of-the-art performance in text summarization, retrieval, and reasoning \cite{chen2025benchmarking}. Yet, their application to structured biomedical datasets, such as PK tables, or high-dimensional functional data, such as electronic health records, network traces, or financial time series, remains limited \cite{TablesMeetLLM}. These domains often exhibit complex structures: biomedical tables contain inconsistent terminologies and units, temporal data involve long-range dependencies, and multidimensional datasets lie on nonlinear manifolds that LLMs do not natively capture \cite{wang2024chainoftableevolvingtablesreasoning}. Consequently, LLMs trained solely on textual corpora struggle to generalize reliably to such domains, resulting in degraded accuracy, F1 Scores, and calibration.

To address these challenges, we propose HySim-LLM, a unified mathematical and computational framework that bridges theoretical guarantees and practical adaptation of LLMs. HySim-LLM integrates functional data analysis, embedding-based similarity metrics, and manifold-aware regularization to enhance the robustness and interpretability of LLMs under domain shift. Building upon our prior AutoPK \cite{sholehrasa2025autopk} system, which applies LLMs to pharmacokinetic table extraction, and its extension, WCPK \cite{ramachandran2023automated}, the HySim-LLM framework generalizes these concepts to a broader theoretical foundation. Specifically, it establishes provable links between embedding similarity, data manifold structure, and the generalization behavior of fine-tuned LLMs.

Our prior work \cite{sholehrasa2025autopk} demonstrated consistent high precision and recall in PK parameter extraction, robust data curation pipelines, and the integration of drug, gene, and adverse-effect information into structured repositories. The HySim-LLM framework advances this foundation by introducing two theoretical results:

\begin{enumerate}
    \item a similarity-weighted fine-tuning bound that quantifies adaptation under embedding divergence; and
    \item a manifold-based denoising theorem that bounds the effect of noisy or off-manifold samples.
\end{enumerate}

Together, these results form a mathematically grounded approach for developing the next generation of generalizable, interpretable, and provably reliable LLMs for biomedical, engineering, and other data-intensive domains.

\section{Related Work}
\subsection*{Domain Adaptation and Generalization Bounds}

The problem of adapting models trained on one distribution to another has been extensively studied in the field of statistical learning theory. Foundational results by Ben-David et al. \cite{ben2010theory} formalized the theory of domain adaptation and introduced generalization bounds based on the $\mathcal{H}$-divergence between source and target distributions. Subsequent extensions incorporated importance weighting and covariate-shift correction to re-balance sample contributions between domains \cite{sugiyama2007direct, cortes2010learning}. More recent work in theory-aware deep learning established generalization bounds for deep networks under smoothness or Lipschitz constraints \cite{neyshabur2018towards, bartlett2017spectrally}.
Our proposed Theorem 1 (Similarity-Weighted Fine-Tuning Bound) builds upon this foundation by introducing embedding-space divergence metrics—such as cosine, Mahalanobis, or Maximum Mean Discrepancy (MMD) distances—into the domain-adaptation bound, providing an interpretable link between semantic similarity and performance guarantees for fine-tuned LLMs.

\subsection*{Embedding Similarity and Transfer in LLMs}

The success of LLMs in few-shot and transfer learning settings has motivated extensive work on embedding-based adaptation. Representation-learning approaches, such as Sentence-BERT \cite{reimers2019sentence}, have demonstrated that well-structured embeddings capture transferable semantics across modalities. Weight-efficient fine-tuning methods, including LoRA \cite{hu2022lora}, LoRA+ \cite{hayou2024lora+}, and AdapterFusion \cite{pfeiffer2021adapterfusion}, focus on parameter efficiency but often lack formal guarantees for adaptation.
In contrast, HySim-LLM unifies embedding similarity with theoretical transfer guarantees, offering provable control over adaptation bias as a function of embedding divergence and source sample size.

\subsection*{Manifold Learning and Denoising}

High-dimensional data in biomedical, veterinary, physical, and engineering domains often lie on low-dimensional manifolds. Classical manifold-learning approaches, such as Isomap \cite{tenenbaum2000global}, Locally Linear Embedding \cite{roweis2000nonlinear}, and Diffusion Maps \cite{coifman2006diffusion}, capture intrinsic structure by estimating neighborhood-preserving embeddings. Modern denoising frameworks, including autoencoders \cite{vincent2008extracting} and diffusion-based representation learning, extend this concept to neural settings.
Our Theorem 2 (Embedding-Based Data Cleaning and Denoising) formalizes this intuition by quantifying how off-manifold samples contribute bounded noise to empirical loss, thereby providing theoretical justification for embedding-space filtering in LLM-based pipelines.

\subsection*{Pharmacokinetic Data Extraction and Curation}

Recent efforts such as AutoPK \cite{sholehrasa2025autopk} and WCPK \cite{ramachandran2023automated} have leveraged LLMs and a rule-based model for PK parameter extraction, schema alignment, and data normalization. These systems demonstrate the promise of LLMs for constructing structured pharmacological knowledge bases, but lack formal guarantees on robustness and generalization. Other related biomedical LLM applications include BioGPT \cite{luo2022biogpt}, PubMedBERT \cite{gu2021domain}, and SciFive \cite{phan2021scifive}, which focus primarily on textual biomedical corpora rather than quantitative table reasoning.
HySim-LLM extends these lines of work by establishing a mathematically grounded framework that unifies LLM adaptation, embedding similarity, and manifold-aware denoising, directly addressing the reliability challenges inherent in PK data extraction and other structured biomedical tasks.

\section{Mathematical Framework}

\subsection{Problem Setup}

Let \(S = \{(x_i, y_i)\}_{i=1}^{n_s}\) be a source dataset, and \(T = \{(x_j, y_j)\}_{j=1}^{n_t}\) a smaller, domain-specific target dataset (e.g., PK tables in AutoPK). Consider a pre-trained LLM with parameters \(\theta_0\) and prediction function \(f_\theta\).

We aim to adapt $\theta_0$ to the target domain using embedding-based similarity metrics while providing provable guarantees for performance (accuracy, F1, or other risk measures).

\subsection{Embedding-Based Similarity Metrics}

Define an embedding function $\mu: \mathcal{X} \to \mathbb{R}^d$, where $d$ is the latent dimension of the model. Let $\mu_T$ denote a representative target embedding (centroid or mixture of prototypes). We define a similarity-based weight for each source example:
\[
\omega_i = \exp(-\alpha \, dist_{\chi}(\mu(x_i), \mu_T)),
\]
where $\alpha > 0$ is a weighting parameter, $dist_{\chi}(\cdot, \cdot)$ is a divergence metric (cosine, Mahalanobis, or kernel using MMD).

The weighted source loss is
\[
L_S^{\omega}(\theta) = \frac{1}{n_s} \sum_{i=1}^{n_s} \omega_i \, \ell(f_{\theta}(x_i), y_i),
\]
where $\ell$ is a bounded loss function (e.g., cross-entropy).

\subsection*{Theorem 1: Similarity-Weighted Fine-Tuning Bound}

We assume that the loss $\ell(f_\theta(x), y)$ is $L$-Lipschitz in $\theta$ and bounded by $B > 0$. Weight constraints are $0 < \omega_i \leq W_{\max}$. Embedding divergence between source and target distributions satisfies $D_{\chi}(p_T \Vert p_S) \leq \delta_\chi$. Also, embedding approximation error is valid for $\|\mu(x_i) - \tilde{\mu}(x_i)\| \leq \epsilon_{\text{embed}}.$  Then, with probability at least $1 - \eta$, we have:
\[
L_T(\theta_\omega) - L_T(\theta_0) \leq C_1 \sqrt{\frac{W_{\max}^2 \, \delta_\chi^2}{n_s}} + C_2 \, \epsilon_{\text{embed}} - C_3 \, \Delta_{\text{opt}}(\theta_\omega, \theta_0) + O\!\left(\frac{1}{\sqrt{n_T}}\right),
\]
where \(\Delta_{\text{opt}}(\theta_\omega, \theta_0) = L_S^{\omega}(\theta_\omega) - L_S^{\omega}(\theta_0) \leq 0\) and \(C_1, C_2, C_3 > 0\) are constants depending on \(L\) and smoothness (activation function) of \(f_\theta\).

\paragraph{Proof (Sketch).}
We decompose the target loss difference:
\[
L_T(\theta_\omega) - L_T(\theta_0)
= \underbrace{L_T(\theta_\omega) - L_S^{\omega}(\theta_\omega)}_{\text{Shift error}}
+ \underbrace{L_S^{\omega}(\theta_\omega) - L_S^{\omega}(\theta_0)}_{\text{Optimization gain}}
+ \underbrace{L_S^{\omega}(\theta_0) - L_T(\theta_0)}_{\text{Reweighting bias}}.
\]

Then the Shift error can be bounded using importance-weight generalization bounds as follows:

\[
|L_T(\theta_\omega) - L_S^{\omega}(\theta_\omega)|
\leq O\!\left(\sqrt{\frac{W_{\max}^2 \, \delta_\chi^2}{n_s}}\right).
\]

Also, the Optimization gain is negative or small under mild convexity or smoothness assumptions. The Reweighting bias arises from embedding mismatch; bounded by \(O\!\left(\frac{1}{\sqrt{n_T}}\right)\). As a result, combining terms yields a bound and a principled selection for \(\alpha\).

To implement the approach, divergence can be estimated using MMD or kernel two-sample tests. A small labeled subset from the target domain should be used to compute the target mean \(\mu_T\). Finally, the LLM can be fine-tuned using the corresponding weights \(\omega_i\).

Noisy or misaligned PK tables, heterogeneous column formats, or mislabeled entries can degrade LLM performance. Embedding-based similarity offers a principled approach to detecting outliers and downweighting or correcting them. This motivation leads to the following theorem.

\subsection{Theorem 2: Embedding-Based Data Cleaning and Denoising}

Assume that true embeddings of `clean' data lie on a low-dimensional manifold \(\mathcal{M} \subset \mathbb{R}^d\). 
Observed embeddings \(\tilde{\mu}(x)\) may contain additive noise \(\varepsilon \sim \mathcal{N}(0, \sigma^2 I_d)\). 
The distance to the low-dimensional manifold can be measured by 
$
d_{\mathcal{M}}(\tilde{\mu}(x)) = \min_{y \in \mathcal{M}} \|\tilde{\mu}(x) - y\|.
$
Let's define a weight function for cleaning by
\[
\omega_i^{\text{clean}} = \exp\!\left(-\beta \, d_{\mathcal{M}}(\tilde{\mu}(x))\right), 
\quad \beta > 0.
\]

Then, with high probability, we have
\[
\frac{1}{n} \sum_{i=1}^{n} \omega_i^{\text{clean}} \, 
\ell(f_{\theta}(x_i), y_i) 
\leq L_{\text{clean}}(\theta) + O(\sigma \sqrt{d}),
\]
where \(L_{\text{clean}}\) is the expected loss over clean manifold-aligned data.

\paragraph{Proof (Sketch).}
First, let's decompose empirical loss into manifold-aligned and off-manifold components. Then we need to weight down off-manifold points using  $\omega_i^{\text{clean}}$. We then use concentration inequalities to bound residual error as a function of noise variance $\sigma^2$ and embedding dimension $d$.

To implement this approach, the underlying manifold can be estimated using methods such as Principal Component Analysis (PCA), autoencoders, or diffusion maps. Embeddings that exceed a specified threshold distance from the manifold $\mathcal{M}$ should be downweighted to reduce their influence. These adjusted weights can then be integrated into the HySim-LLM fine-tuning process or within the AutoPK extraction pipeline.

\subsection{Integration into HySim-LLM}

For integration into HySim-LLM, weighted fine-tuning can be performed using the Theorem~1 weights to facilitate target adaptation. 
Data cleaning should employ the Theorem~2 weights to identify and either remove or downweight noisy PK entries. 
A hybrid loss function can then be constructed by combining the two sets of weights, either multiplicatively or additively, such that 
$
\omega_i^{\text{hybrid}} = \omega_i \, \omega_i^{\text{clean}}.
$
The end-to-end algorithm fine-tunes the LLM using a mixture of source data, cleaned data, and similarity-weighted target examples. 
Finally, evaluation involves measuring F1 or accuracy improvements and empirically verifying the theoretical performance bounds.

\section{Algorithmic Implementation}

\subsection*{Algorithm 1: HySim-LLM Weighted Fine-Tuning}

\begin{algorithm}[H]
\caption{HySim-LLM Weighted Fine-Tuning}
\begin{algorithmic}[1]
\REQUIRE Source dataset $S = \{(x_i, y_i)\}$, Target dataset $T = \{(x_j, y_j)\}$, Pre-trained LLM $f(\cdot; \theta_0)$, Embedding model $\mu(\cdot)$, Parameters $\alpha, \beta$
\ENSURE Fine-tuned parameters $\hat{\theta}$

\STATE Compute embeddings $\mu(x_i)$ for $x_i \in S$, $\mu(x_j)$ for $x_j \in T$.
\STATE Compute target centroid $\mu_T = \frac{1}{|T|} \sum_j \mu(x_j)$.
\STATE For each source sample, compute $\omega_i = \exp(-\alpha\, dist_{\chi}(\mu(x_i),\mu_T))$.
\STATE Estimate manifold $\mathcal{M}$ via PCA or autoencoder.
\STATE Compute $d_\mathcal{M}(\mu(x_i))$ and $\omega_i^{\text{clean}} = \exp(-\beta\, d_\mathcal{M}(\mu(x_i)))$.
\STATE Combine weights: $w_i^{\text{total}} = \omega_i \cdot \omega_i^{\text{clean}}$.
\STATE Minimize $L(\theta) = \sum_i w_i^{\text{total}} \, \ell(f(x_i; \theta), y_i)$ using AdamW or L-BFGS with learning-rate warm-up..
\STATE Evaluate F1, Accuracy, and Expected Calibration Error (ECE) on target validation data.
\end{algorithmic}
\end{algorithm}

\subsection*{Algorithm 2: AutoPK Data Extraction and Cleaning}

\begin{algorithm}[H]
\caption{AutoPK Data Extraction and Cleaning}
\begin{algorithmic}[1]
\REQUIRE Raw pharmacokinetic tables (CSV, PDF, or HTML)
\ENSURE Clean, normalized PK table ready for model input

\STATE Parse schema using LLM templates (e.g., map \textit{animal} $\rightarrow$ \textit{species tag}, \textit{compound} $\rightarrow$ \textit{drug name}, \textit{parameters} $\rightarrow$ \textit{Cmax, AUC, t$\frac{1}{2}$, etc.}).

\STATE Detect units with a regular-expression library and normalize to canonical SI units using learned conversion embeddings.
\STATE Compute $\mu(x)$ for each row vectorized as [Cmax, AUC, t$\frac{1}{2}$, CL, Vd].
\STATE Reject or downweight rows with $d_\mathcal{M}(\mu(x)) > \tau$, where $\tau = \text{mean} + 2 \cdot \text{std}$ of in-manifold distances.
\STATE Feed cleaned, weighted rows to HySim-LLM fine-tuning loop.
\end{algorithmic}
\end{algorithm}

% \section{Results}

% \section{Discussions}

\section{Future Work}
\subsection{AutoPK dataset}
\begin{figure}[htbp]
  \centering
    \vspace{-2em}
  \begin{subfigure}[t]{0.48\textwidth}
    \centering
    \includegraphics[height=0.4\textheight]{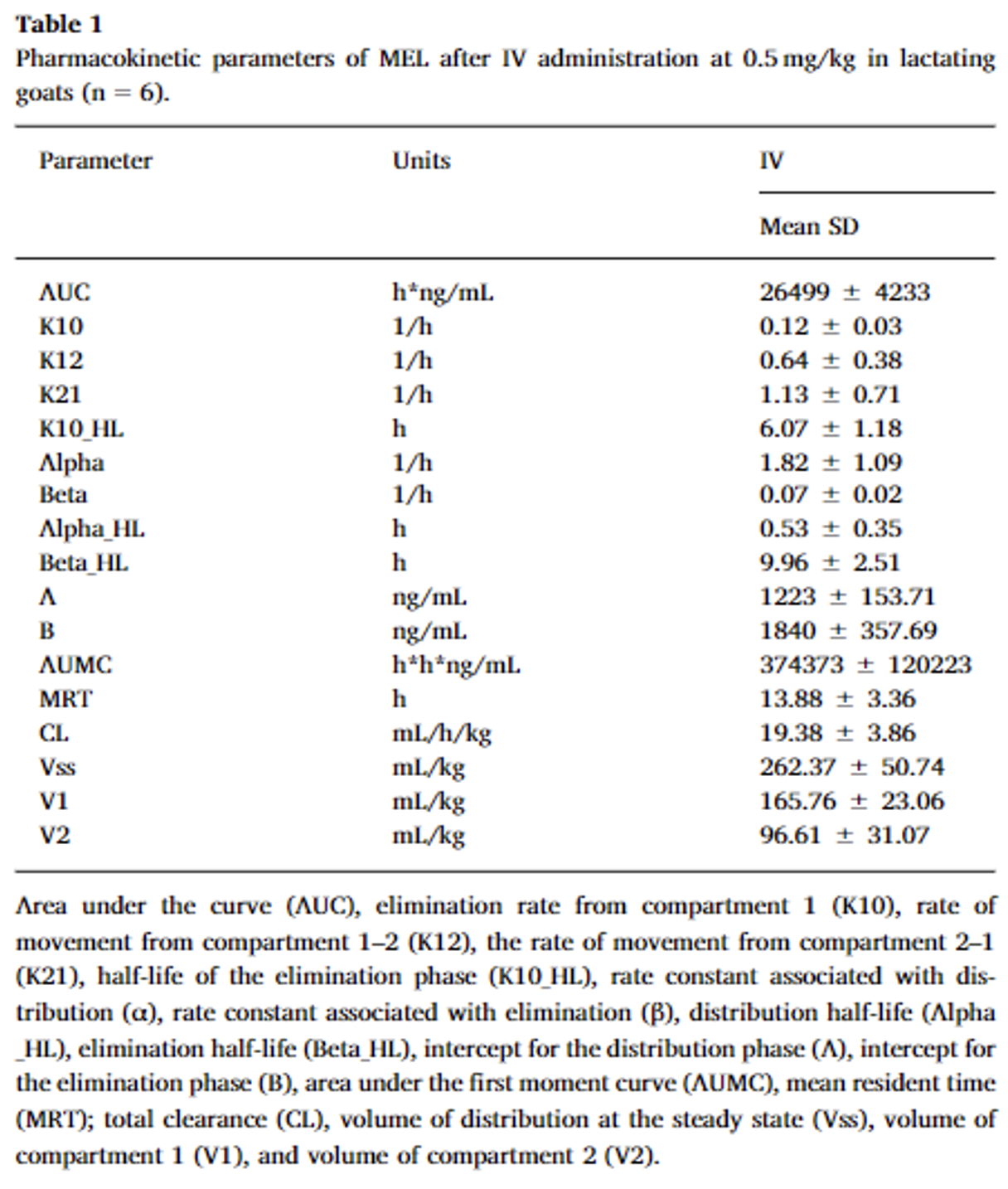}
    \caption{Original PK table as published in a scientific article. 
    The table presents PK parameters (e.g., AUC, K10, K12) with corresponding units and summary statistics (Mean $\pm$ SD). 
    Such tables often include complex multi-row headers and embedded textual notes.}
  \end{subfigure}
  \hfill
  \begin{subfigure}[t]{0.48\textwidth}
    \centering
    \includegraphics[height=0.4\textheight]{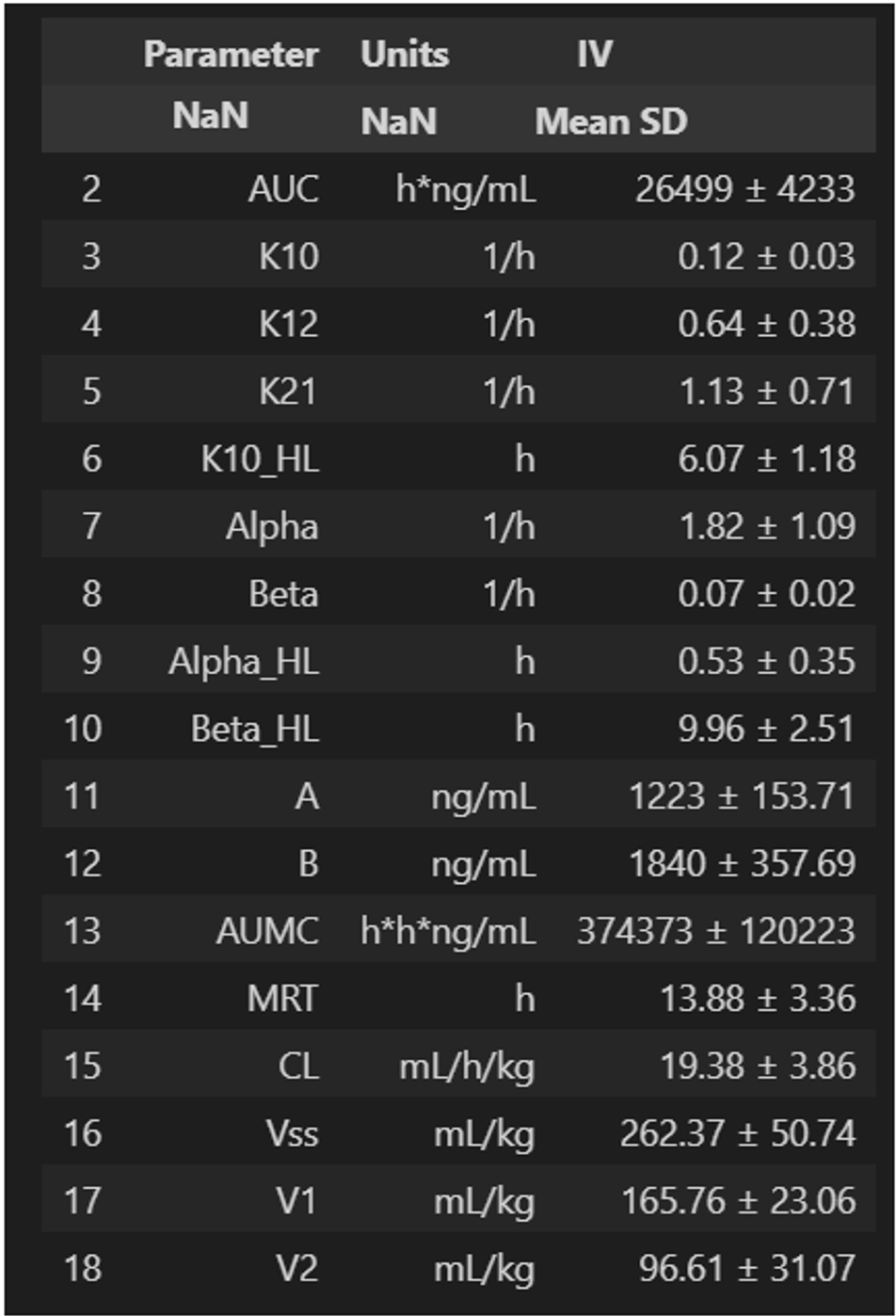}
    \caption{The extracted and normalized version of the same PK table. Each parameter, unit, and value is parsed and structured into machine-readable fields for downstream data analysis and modeling.}
  \end{subfigure}
  \caption{Comparison between the raw published PK table and its automatically extracted structured representation from the AutoPK dataset.  This illustrates the transformation from unstructured scientific table formats into standardized, analysis-ready tabular data used for dataset curation and model evaluation.}
  \label{fig:input_extracted}
\end{figure}

We utilized the real-world PK table dataset introduced in our prior work \cite{sholehrasa2025autopk}. This dataset comprises scientific tables and their corresponding textual context, including captions, footnotes, and the title and abstract of the associated scientific articles. A summary of its key statistics is provided in Table~\ref{tab:dataset_stats}, which was used to evaluate our prior work using the 605 annotated tables. An illustrative example of the table extraction process is shown in Figure \ref{fig:input_extracted}.

The dataset was originally collected using a PK-specific web crawler \cite{ramachandran2023automated} that retrieved 1,522 tables containing PK data from 1,088 XML-formatted full-text scientific articles. It then extracted relevant table information through automated XML parsing and normalization. From these articles, the title, abstract, and all table-related content—including data cells, captions, and footnotes—are parsed by using relevant XML tags. In this work, we employ the same dataset for evaluation and fine-tuning purposes. Furthermore, we plan to extend the dataset by applying the same data-gathering and preprocessing pipeline to additional scientific publications, thereby increasing coverage across species, study types, and experimental conditions. Future work will focus on fine-tuning LLMs on the AutoPK dataset using the HySim-LLM to enhance generalization across heterogeneous PK table domains while mitigating noise and adaptation bias.

\begin{table}[ht]
\vspace{-1em}
\centering
\caption{Descriptive statistics of the AutoPK dataset, covering average table dimensions, structural characteristics, and counts of PK parameter variants \cite{sholehrasa2025autopk}.}
\begin{tabular}{>{\raggedright\arraybackslash}p{9cm} c}
\hline
\textbf{Statistic} & \textbf{Values} \\
\hline
\#Tables & 605 \\
Avg \#rows/cols/multi-header-rows input tables & 8.63 / 5.43 / 2.35 \\
Avg \#rows/cols output tables & 21.56 / 8.00 \\
Unique HL / AUC / CL variants & 338 / 602 / 370 \\
Unique MRT / CMAX / TMAX variants & 61 / 161 / 74 \\
Single/multi-header/block-structured table types & 62\% / 26\% / 12\% \\
\hline
\end{tabular}
\label{tab:dataset_stats}

\end{table}

\subsection{Hybrid Mechanistic-LLM Models}
Future work will explore coupling HySim-LLM with mechanistic pharmacokinetic models, such as compartmental ODE systems, to enable hybrid inference. Learned embeddings can serve as priors or regularizers for parameter estimation, linking data-driven adaptation with physiologically grounded dynamics. This integration aims to enhance interpretability, improve parameter stability, and unify empirical and mechanistic modeling approaches within pharmacokinetic analysis.
\subsection{Broader Applications}
Beyond PK, HySim-LLM can be extended to diverse biomedical domains that involve structured quantitative data, such as pharmacovigilance reports, therapeutic response profiles \cite{golmohammadi2025comprehensive}, toxicological assays, and clinical outcome datasets \cite{sholehrasa2025predictive, xu2019making}. In clinical pharmacology, the framework could support dose optimization, therapeutic drug monitoring, and individualized treatment modeling by aligning patient-specific PK profiles with reference manifolds. In toxicology and systems biology, embedding-weighted adaptation may improve cross-species prediction of exposure or clearance rates. In genomics and transcriptomics, manifold-aware denoising can enhance the extraction of regulatory patterns from noisy, high-dimensional omics data. These directions provide natural testbeds for validating the theoretical guarantees of HySim-LLM across biomedical research pipelines where data heterogeneity and noise remain key challenges.
\section{Conclusion}
In this work, we introduced HySim-LLM, a unified mathematical and computational framework that provides theoretical guarantees for adapting LLMs to structured and domain-specific data. By formulating similarity-weighted fine-tuning bounds and a manifold-based denoising theorem, we established provable links between embedding similarity, data geometry, and generalization performance under domain shift. These results bridge theoretical learning guarantees with practical implementation through the HySim-LLM pipeline, which integrates embedding-based weighting and manifold-aware data cleaning into an end-to-end fine-tuning process.

\bibliographystyle{unsrt}
\bibliography{references}

\end{document}